# A Deep Neural Framework for Image Caption Generation Using GRU-Based Attention Mechanism

Rashid khan[a], M Shujah Islam[a], Khadija Kanwal[a], Mansoor Iqbal, Md. Imran Hossain[a] & Zhongfu Ye[a]*

[a] *National Engineering Laboratory for Speech and Language Information Processing, University of Science and Technology of China, Hefei, 230026, Anhui, China*
**E-mail:** {rashidkhan, mshujahislam, khadijakanwal, man2017, imranpost *1@mail.ustc.edu.cn* & *yezf@ustc.edu.cn*

**Abstract**

Image captioning is a fast-growing research field of computer vision and natural language processing that involves creating text explanations for images. This study aims to develop a system that uses a pre-trained convolutional neural network (CNN) to extract features from an image, integrates the features with an attention mechanism, and creates captions using a recurrent neural network (RNN). To encode an image into a feature vector as graphical attributes, we employed multiple pre-trained convolutional neural networks. Following that, a language model known as GRU was chosen as the decoder to construct the descriptive sentence. In order to increase performance, we merged the Bahdanau attention model with GRU to allow learning to be focused on a specific portion of the image. On the MSCOCO dataset, the experimental results achieve competitive performance against state-of-the-art approaches.

## 1. Introduction

Image caption generation is a challenging artificial intelligence task in which a textual explanation for a given image must be generated. It needs both computer vision approaches to understand the image's content and a language model from the field of natural language processing to convert the image's understanding into words in the right sequence. with the development of deep learning in computer vision, the twenty-first century, or the Artificial Intelligence Era, began. The main aim of the data scientists was to extract visual features from images, identify them, and detect objects. in recent years, Natural Language Processing algorithms have played an increasingly important role in deep learning research, allowing computers to learn from the world of words, interpret them, classify them, and extract features. Natural language processing (NLP), on the other hand, has spawned a slew of applications ranging from basic text classification to fully automated natural language chatbots. In the Deep Learning domain, it has been a critical and fundamental challenge. Captioning images has a wide range of applications, including (i) transcribing scenes for people who are blind [1][2], (ii) classifying videos and photographs based on various situations [3], (iii) image-based search engines for better results, [4] (iv) visual question answering [5], and (v) context comprehension [6].

To generate the corresponding sentence for one given image, latest research on caption generation, such as image captioning [7], relies on an encoder-decoder framework. Different neural network architectures are employed as the encoder due to the different behavior and features of the source, like convolutional neural networks (CNNs) for images and recurrent neural networks (RNNs) for sequential data also including source code and natural language. The attention technique evaluates the significance of the encoder's hidden states based on all previously generated words in the target sentence at each time step. The attention mechanism, on the other hand, works in a sequential manner and lacks global modeling capabilities. A review network [8] was proposed to overcome this flaw, with review steps located here between encoder and the decoder. As a result, more compact, abstractive, and global annotation vectors are generated, which have been shown to assist the sentence generation process further in.

We introduced a CNN and GRU-based attention mechanism for automatic image captioning in this research work. The suggested frame work was designed with one encoder-decoder frame work. As the encoder, we used various pre-trained convolutional neural networks to encode an image into a feature vector as graphical characteristics. Subsequently, to create the descriptive sentence, a language model called GRU was chosen as the decoder. However, we combined the Bahdanau attention model with GRU to allow learning to be focused on a specific portion of the image in order to improve the performance.

The following are the main contributions of our paper:

- For image caption generation, we examined into the encoder-decoder framework. The ENCODER would use a pre-trained Convolutional Neural Network to encode the image, and the DECODER would use a Recurrent Neural Network to create each word of the caption iteratively (RNN).
- The model's performance was compared to that of four pre-trained CNNs: InceptionV3, DenseNet169, ResNet101, and VGG16.
- We employ the RNN (GRU) with a soft attention as the decoder, which effectively focuses the attention over a certain part of an image to predict the next sentences.
- In our image captioning approach, we apply an attention mechanism that can focus on the important elements of the image and define fine-grained captions.
- Finally, we utilize the public MS COCO dataset [9] to quantitatively validate the research paper's utility in image caption generation.

The rest of this research paper is structured in the following manner. Section 2 we start by going through the previous work and an overview of our framework. Following that, each module of our technique is presented in depth. In section 3, we describe our proposed framework in detail. The results of our approach are shown in Section 4 as an evaluation of our methodology. In section 5, we'll go through the future directions in depth and how they may be further explored.

## 2. Literature Review

In related work we enhance relevant information on prior study on image caption generation and attention. Several approaches for generating image descriptions have recently been presented. Automatic image captioning generation has emerged as a promising research area in recent years, because to advances in deep neural network models for Computer Vision (CV) and Natural Language Processing (NLP). In general, there are three types of image captioning modeling techniques: neural-based approaches [10] [11] [12], attention-based strategies [13] [14] [15] [16], and RL-based methods framework [17] [18]. Attention-based approaches have recently gained popularity and have been shown to be more successful than neural-based methods. When guessing each word in the caption, attention-based techniques tend to focus on certain locations in the image.

Deep neural networks (DNNs) were initially proposed for caption generation [19]. They suggested extracting characteristics from images using convolutional neural networks (CNNs) to produce captions. A popular captioning technique involves integrating CNN and RNN, with CNN extracting image features and RNN framework generating the language model [20]. for example, presented an end-to-end network made up of a CNN and an RNN. Given the CNN feature of the training image at the starting time step, the model is trained to maximize the likelihood of the target sentence. The image's CNN feature is provided into the multimodal layer after the recurrent layer rather than at the beginning in the proposed m-RNN model [21]. Some are comparable instances of similar work [22] [23] that utilizes CNN and RNN to generate descriptions.

Visual attention has been shown to be an efficient approach for image captioning generation [24] [25]. When developing the target language, these attention-based captioning models may learn where to focus in the image. They may learn the distribution of spatial attention during the last convolutional layer of the CNN [26], or they learn the distribution of semantic attention from visual characteristics learned from social media images [27]. Whereas these methodologies demonstrate the efficiency of the attention mechanism, they do not investigate the contextual information in the encoding sequence. Our attention layer is distinct in that it is structured in a sequential order, with each hidden state of an encoding stage contributing to the formation of decoding words.

An Attention mechanism can be used to improve the contextual aspect of natural language sequences. The use of attention to describe image content is consistent with human understanding [24]. The evaluation matrix and the accuracy of attention to an image have a significant correlation. Even still, the measure to which attention accuracy is congruent with human perceptions needs to be increased [28]. The attention area captioning model is made up of three parts: image regions, word captions, and the NLP natural language framework (RNN). The MS COCO dataset is typically used to test the trained system [29].

In artificial intelligence, developing a caption that accurately represents an image is essential [ 30]. One of the procedures in image captioning is extracting coherent characteristics of an image using an image-based framework. The extracted features of the image-based model are used to describe an image in NL. We suggest to create an image captioning model that employs an RNN (GRU) with a soft attention decoder to predict the future sentences by selectively focusing attention over a specific part of an image. We used cutting-edge architecture to evaluate the model's performance to that of four pre-trained CNNs: Inception V3, DenseNet169, ResNet101, and VGG16. The Attention layer is used to make the image's caption more sensible.

## 3. Caption Generation using GRU-based attention Network

Extracting visual information and expressing it in a grammatically accurate natural language sentence are the two main components of automatically generating natural language sentences that describe an image. For image captioning, Fig. 1, shows a simple Encoder-Decoder deep learning-based captioning infrastructure.

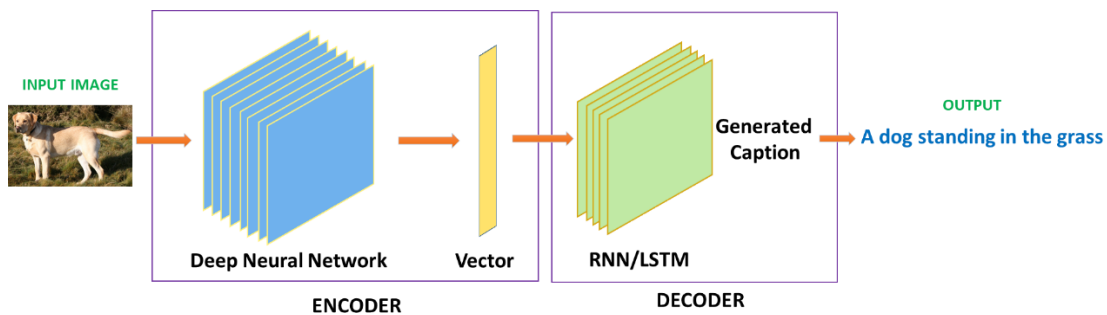

**Fig.1** This is an image captioning model's overall encoder-decoder structure. The image is encoded into a feature vector by a deep neural network. The input vector is used by the language model to construct a sentence that defines the image.

As previously stated, our approach is significantly influenced by the work of Xu et al. [31]. Their performance has improved as a result of our efforts. The *CNN-Encoder, attention mechanism*, and *RNN-decoder* are the three major components of our proposed framework. They work in sequence, with the images with captions as the CNN-encoder's input. The attention technique and the LSTM work in conjunction to generate captions for the input image, and the output of this is passed to them. The objects and features in the image are retrieved using a convolutional neural network, and then we require a network to construct a meaningful sentence using the information we have.

Our proposed model generates a caption $q$ encoded as a sequence of $1-of-K$ encoded words from a single input image.

$$q = \{q_1, q_2 \ldots, q_C\}, qi \in R^K \qquad \text{Equation: (1)}$$

Where $K$ denotes the vocabulary size and C denotes the caption length. A pre-trained CNN as encoder extracts features from images, an attention mechanism weights the image features, and an RNN as decoder provides captions to represent the weighted image features. The overall graphical representation of our framework is shown in Fig. 2.

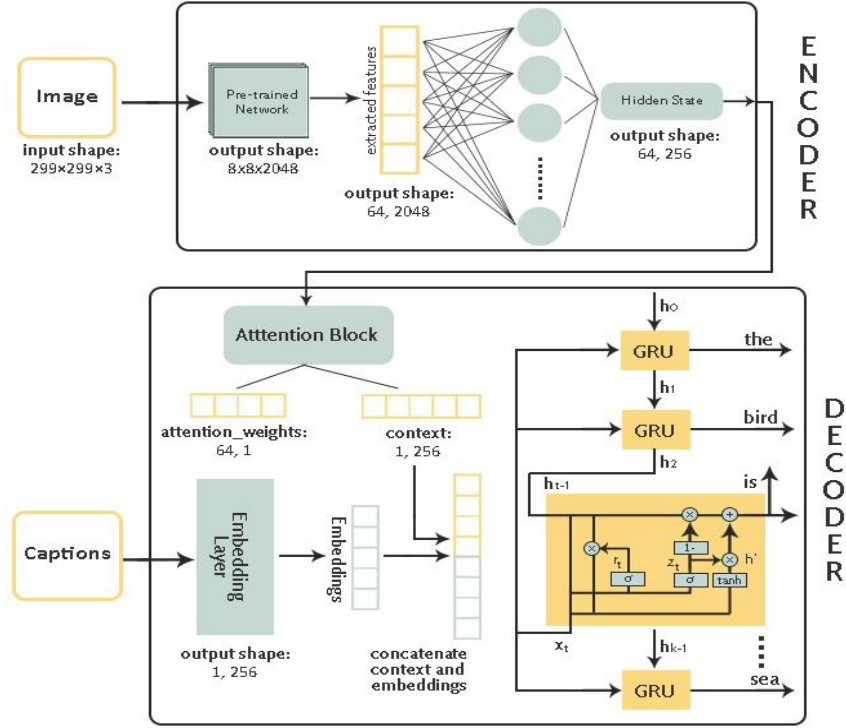

**Fig. 2**   Figure showing visual representation of approach. It is based on Xu et al [31].'s architecture proposal. The input image is sent to the encoder, which then transfers the output of the convolutional layers to the GRU with attention mechanism.

### 3.1   Convolutional Neural Network (Encoder)

A CNN pre-trained for an image classification task is commonly used as the encoder in the encoder-decoder framework for image captioning to extract the global representation and sub region representations of the input image. A fully connected layer's output is usually the global representation, while a convolutional layer's output is usually the sub region representation. As an encoder, we employed a convolutional neural network. A common feed-forward neural network is a convolutional neural network. Convolutions and element-wise summations are performed by each layer as part of an affine operator:

$$f(x) = \sum_{k=1}^{K} g_k * x + b_k \qquad \text{Equation.} \quad (2)$$

We need to convert each image into a fixed-size vector that can then be provided as input to the neural network in this research study. We chose below four pre-trained architectures for this purpose.

a) **Inception:** Inception-v3 is a $48-layer$ deep pre-trained convolutional neural network model. The network accepts a 299-by-299 image as input and outputs an $8*8*2048$ feature vector.

b) **DenseNet169**: DenseNet-169 is a $169-layer$ model that has been pre-trained. It has fewer parameters than other approaches, and the architecture effectively tackles the vanishing gradient problem. The network accepts a 224-by-224 image as input and outputs a $7*7*1664$ feature vector.

c) **ResNet101:** ResNet-101 is a $101-layer$ deep convolutional neural network. The network learns rich feature representations for a wide range of images because to the vast number of layers. The network's image input size is $224 \times 224$ pixels. It generates a $7*7*2048$ feature vector.

d) **VGG16:** The VGG pre-trained model was released by researchers from the Oxford Visual Geometry Group, who took part in the ILSVRC challenge. The model expects color input photos to be rescaled to $224 \times 224$ squares by default. It generates a $7*7*512$ feature vector.

These networks were trained to do image classification on 1000 different classes of images using the ImageNet dataset. Our goal isn't to classify the image, but rather to obtain a fixed-length informative vector for each image. As a result, we eliminated the model's last softmax layer and extracted a fixed length vector for each image. These extracted characteristics are sent to the encoder, which creates the hidden state by using a fully connected layer.

### 3.2 Recurrent Neural Networks (RNN)

While CNNs perform at signal processing, they tend to describe patterns in sequences. The outputs of recurrent neural networks (RNNs) are fed back into the input, similar to feedforward neural networks. Throughout iterations of this feedback loop, the networks preserve a hidden state that allows them to change their behavior. RNNs are regarded state-of-the-art in machine translation and other jobs involving text generation because they acquire human grammatical patterns quickly. In this research paper, we use a Gated recurrent unit (GRU), a sophisticated variant on the RNN approach that produces a caption by creating one word at each time step based on a context vector, the prior hidden state, and previously created words.

The most likely description of an image is obtained in the encoder-decoder approach by maximizing the log-likelihood function of the expression $E$, considering the related image $I$ and the model parameters $\boldsymbol{\theta}$.

$$\theta^* = \arg\max \sum (I, E) \log p(S|I; \theta) \qquad \text{Equation.} \quad (3)$$

Because $S$ can represent any length of sentence, a chain rule is commonly employed to characterize the joint probability over $E_1, E_2 \cdots, E_N$.

$$\log p(E|I) = \sum_{t=0}^{N} \log p(E_t|I, E_0, \ldots, E_{t-1}) \qquad \text{Equation.} \quad (4)$$

For the sake of clarity, the dependency on $\theta$ is excluded. The network training is represented by the pair of $(E, I)$, and we use the Adam optimizer to maximize the sum of the log likelihood functions across the full training set. A recurrent neural network is used to represent the likelihood $\log p(E_t | I, E_0, E_1 \cdots, E_{t-}$

1), where there is a variable number of words that we define up to $t-1$. Using a fixed sized hidden vector, recurrent neural networks (RNN) give a seamless technique to execute conditioning on prior variables.

$$p(E_t|I, E_0, E_1 ...., E_{t-1}) \approx p(E_t|I, h_t) \qquad \text{Equation.} \quad (5)$$

As a result, at step $t$, a simple vector replaces the complex conditioning on a variable number of nodes ($h_t$). After the new input $x_t$, the RNN's hidden state (latent memory) $h_t$ is updated with the nonlinear function $f$.

$$h_{t+1} = f(h_t, x_t) \qquad \text{Equation.} \quad (6)$$

The capacity of $f$ in Equation (6) to deal with vanishing difficulties and exploding gradients, which are the most typical problems in the development and training of RNN, determines the value of $f$. Given the inputs $X_t, h_{t-1}$, the $GRU$ updates for time step $t$. To begin, we use the formula to calculate the update gate $z_t$ for time step $t$:

$$z_t = \sigma(W_z x_t + U_z h_{t-1}) \qquad \text{Equation.} \quad (7)$$

When $X_t$ is linked through into network unit, its own weight $W_z$ is multiplied. The same is true for $h_{t-1}$, which stores data from earlier $t-1$ units and is multiplied by their own weight $U_z$. Both results are combined together, and the result is squashed between $0$ $and$ $1$ using a sigmoid activation function. The update gate aids the model in determining how much previous data (from earlier time steps) should be passed on to the future. This is extremely useful since the model can choose to duplicate all of the data from the past, eliminating the risk of disappearing gradients. The reset gate is utilized by the framework to determine how much information from the past should be forgotten. We use the following formula to compute it:

$$r_t = \sigma(W_r x_t + U_r h_{t-1}) \qquad \text{Equation.} \quad (8)$$

We create a new memory content that will store relevant information from the past using the reset gate. It is calculated as follows:

$$h_t = \tanh(Wxt + r_t \odot U h_{t-1}) \qquad \text{Equation.} \quad (9)$$

This multiplies the input $x_t$ by a weight $W$ and the input $h_{t-1}$ by a weight $U$. Calculate the Hadamard (element-by-element) product between $r\_t$ and $U h_{t-1}$. This will determine what time steps should be removed from the preceding ones. Add the results together and use the $tanh$ nonlinear activation function. The network's final step is to calculate the $h_t$ vector, which contains information for the current unit and sends it down to the network. The update gate is required to accomplish this. It determines what should be collected from the present memory content $h't$ and what should be collected from the previous stages. $h_{t-1}$. This is how it's done:

$$h_t = z_t \odot h_{t-1} + (1 - z_t) \odot h'_t \qquad \text{Equation.} \quad (10)$$

This will execute element-wise multiplication on the updates gates $z_t$ and $h_{t-1}$, as well as $(1 - z_t)$ and $h't$, and total the results. Below Fig. 3, is a visual representation of GRU.

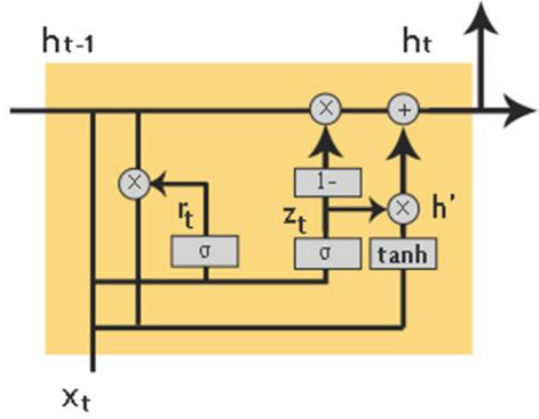

**Fig. 3** Visual representation of Gated Reccurrent Unit (GRU)

We want the representation for words to be such that the vectors for words with similar meanings or contexts are close to one other in the vector space. The word2vec algorithm, which turns a word into a vector, is likely the most used. A corpus of texts relating to the domain in which we are engaged is used to train word2vec. Word2vec trains a system that can anticipate the surrounding words of a target text in order to calculate the word vectors. Surrounding words are described as words that appear from both side of a given word in a small context window of a particular size. These three sentences will serve as our corpus. (1) *This research paper is about deep learning and computer vision*. (2) *We love deep learning*. (3) *We love computer vision.*

The challenge is to predict the context terms *"learning*," *"and,"* and *"vision"* from the first phrase and *"we," "love,"* and *"vision"* from the last sentence given the word *"computer."* As a result, the goal of training is to maximize the log probabilities of these context terms given the word "computer." The formulation is:

$$\text{Objective} = \text{Maximize} \sum_{t=1}^{T} \sum_{-m \geq j \geq m} \log P(E_{wt} + j | E_{wt}) \quad \text{Equation.} \quad (11)$$

Where m is the size of the context window and *t* is the length of a corpus. The similarities or inner product between the context word vector as well as the center word vector is used to represent $P(E_{wt+j}|E_{wt})$. When a word appears in context and when it is the central word, it has two vectors associated with it, denoted by *R and S*, accordingly. As a result, $P(E_{wt+j}|E_{wt})$ is defined as:

$$\frac{e^{R^T_W{t+j}}}{\sum_{i=1}^{M} e^{R_i^T s_{wt}}} \quad \text{Equation.} \quad (12)$$

The denominator is the normalization term, which compares the similarity of the central word vector to a context vectors of every other word in the lexicon, resulting in a probability of one. The vector representation for a word is then chosen as the center vector, as with GRU.

## 3.3 Attention models

We apply the Bahdanau attention mechanism to better isolate the image content, which has been widely used to tackle the challenge of image categorization since it eliminates the need to process every pixel in an image. Instead of taking features from the entire image, the salient portion of the image is determined at each step and input into the RNN. The algorithm uses the image to create a focused view and predicts the term that is relevant to that location. On the basis of previously generated words, the location where attention is directed must be determined. Otherwise, new words formed within the region may be coherent, although not in the description is generated. Let's look at the mechanism proposed by Bahdanau now. To begin, calculate the $e_{jt}$ score.

$$e_{jt} = f_{ATT}(E_{t-1}, h_j) \quad \text{Equation. (13)}$$

$e_{jt}$ is a score that indicates how essential the $j^{th}$ pixel of an image is at every time step $t$ of the decoder. The prior state of the decoder is $S_{t-1}$, while the current state of the encoder is $h_j$. $f_{ATT}$ is a basic feed forward neural network that sums $S_{t-1}$ and $h_j$ from the fully connected layer, then passes it through a non-linear function $tanh$ before returning to the fully connected layer.

$$e_{jt} = FC(tanh(FC(E_{t-1}) + FC(h_j))) \quad \text{Equation. (14)}$$

We use softmax to get the probability distribution.

$$\alpha_{jt} = softmax(e_{jt}, axis=1) \quad \text{Equation. (15)}$$

Softmax is normally applied to the last axis, however since the shape of the score is (batch_size, max_length, hidden_size), we want to apply it to the 1st axis. The maximum length of our input is max length. Softmax should be applied to that axis because we're aiming to allocate a weight to each input. Now that we have the input, we must feed the decoder a weighted sum combination of the input.

$$c_t = \sum_{j=1}^{T} a_{jt} h_j \quad such\ that\ \sum_{j=1}^{T_x} a_{jt} = 1\ and\ a_{ij} \geq 0 \quad \text{Equation. (16)}$$

The context vector (weighted total of the input) that will be sent to RNN is $c_t$.

$$E_t = RNN(S_{t-1}, e(y'_{t-1}), c_t) \quad \text{Equation. (17)}$$

The previous state of the decoder is $S_{t-1}$, and the previous predicted word is $e(y'_{t-1})$.

## 4. Experimental Setup and Results

The objective of image captioning is to create a natural sentence that describes the visual information of a single image. The most prominent MSCOCO dataset [9] is used in this work to demonstrate the usefulness of our proposed framework.

## 4.1 Data and Metrics

The MSCOCO dataset [9], which contains 82,783, 40,504, and 40,775 images for training, validation, and test, is the largest benchmark dataset for the image captioning task. Because most images feature many objects in the context of complicated situations, this dataset is difficult to analyze. Each image in this dataset has five captions different ground truth that have been annotated by humans, as seen in Fig. 4. We use the same data split as in [23] for offline assessment, which consists 5,000 photos for validation, 5,000 images for test, and 113,287 images for training.

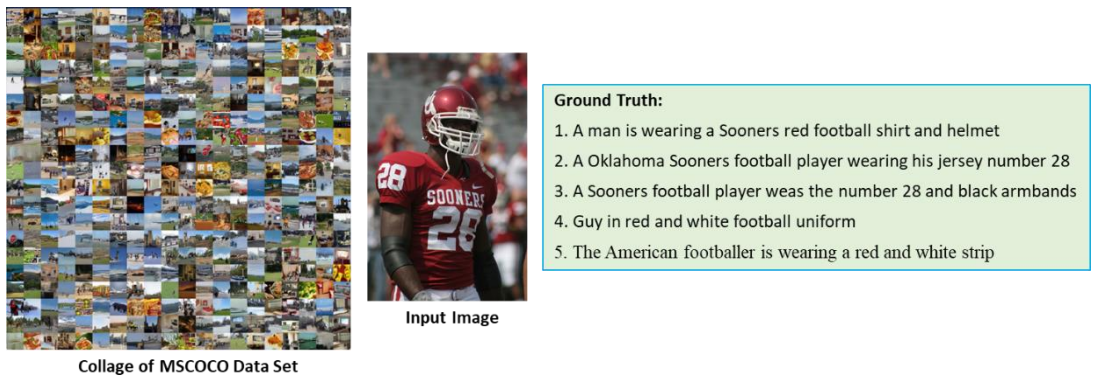

**Fig.4**     Example of Microsoft Common Objects in COntext (MS-COCO) dataset.

We combine the testing set with the training set to create a larger training set for online evaluation on the MSCOCO evaluation server. We remove all non-alphabetic characters from the captions, convert all letters to lowercase, then tokenize the captions with white space. As a result, a vocabulary of 9,487 terms has been created.

## 4.2 Evaluation measurements

Four commonly used evaluation metrics, namely BLEU1-4 [32], Meteor [33], Rouge-L [34], and CIDEr [35], are utilized to evaluate the quality of generated sentences utilizing the publicly accessible MSCOCO tools to quantitatively analyze the performance of our proposed methods. All of these metrics are used to compare the consistency of n-grams in generated and reference sentences. To make accurate comparisons with existing image captioning approaches.

## 4.3 Quantitative Results

We present quantitative results in this section to demonstrate the efficacy of the proposed strategy. We compare the suggested technique to seven state-of-the-art models in a multi comparison. In the first-time step of the LSTM-based language model, NIC injects image features derived from the fully connected layer of a deep CNN. The results presented in [36] are cited directly. Soft-Att selects some regional representations from the deep CNN's final convolutional layer and uses an LSTM-based language model to decode each word at each timestep depending on the representations selected [31]. MSM incorporates inter-attribute correlations into a multiple instance learning approach and investigates several approaches of injecting detected characteristics and image representations into an LSTM-based language framework [2]. Attribute-driven Using a CNN-RNN architecture and using the visual attention method for attribute detector, attention model the co-occurrence dependencies among attributes [37]. NBT architecture for visually grounded image captioning that generates free-form natural language descriptions while clearly localizing things in the image. [38]. The relationship between the objects/regions in an image is modelled with a GNN in visual context-aware attention, which considers the visual relationship between regions of interest for improved representation of the visual content in the image [39]. The elegant views of what kind of visual relationships could be built between objects, and how to nicely leverage such visual relationships to learn more informative and relation-aware region representations come from GCN-use LSTM's of visual relationships for enriching region-level representations and eventually enhancing and leads to the elegant views of what kind of visual relationships could be built between objects and how to nicely leverage such visual relationships to learn more informative and relation-aware region representations [40].

Table 1 Comparison of the proposed GRU attention-based model with various baseline approaches on the MSCOCO dataset

| **Experimental Results of state-of-the-art methods on MS-COCO** | | | | | | | |
|---|---|---|---|---|---|---|---|
| **MODEL** | **BLEU-1** | **BLEU-2** | **BLEU-3** | **BLEU-4** | **Rouge** | **CIDER** | **METEOR** |
| Google NIC [36] | 0.67 | 0.45 | 0.30 | 0.20 | -- | -- | -- |
| Soft Attention [31] | 0.71 | 0.49 | 0.34 | 0.24 | -- | -- | 0.24 |
| MSM [2] | 0.73 | 0.57 | .043 | 0.33 | 0.54 | 1.02 | 0.25 |
| Attribute-driven Attention [37] | 0.74 | 0.56 | 0.44 | -- | 0.55 | 1.104 | -- |
| NBT [38] | 0.75 | -- | 0.34 | -- | -- | 1.107 | 0.27 |
| Context-aware attention [39] | 0.76 | 0.60 | 0.46 | 0.36 | 0.56 | 1.103 | 0.28 |
| GCN-LSTM [40] | 0.77 | -- | -- | 0.36 | 0.57 | 1.107 | 0.28 |
| **Performance of our proposed GRU attention-based models** | | | | | | | |
| **MODEL** | **BLEU-1** | **BLEU-2** | **BLEU-3** | **BLEU-4** | **Rouge** | **CIDER** | **METEOR** |
| Inception V3 | **0.78** | **0.57** | **0.44** | 0.36 | **0.59** | 1.105 | 0.27 |
| VGG16 | 0.74 | **0.57** | **0.44** | 0.33 | 0.56 | **1.109** | 0.26 |
| DenseNet169 | 0.74 | 0.56 | 0.43 | 0.36 | **0.58** | 1.103 | 0.27 |
| ResNet101 | 0.75 | 0.56 | **0.44** | **0.37** | **0.59** | 1.104 | **0.29** |

So far, the outcomes of our research have been favorable. The results of our model's quantitative analysis were given using four metrics: BLEU, Rouge, CIDER, and Meteor. Other state-of-the-art approaches were compared to the performance of different pre-trained models for encoder architecture. The bold number represents the top results for that measure whereas those with a dash (--) are unavailable. Table 1 shows that Inception outperformed the other models followed by ResNet101.

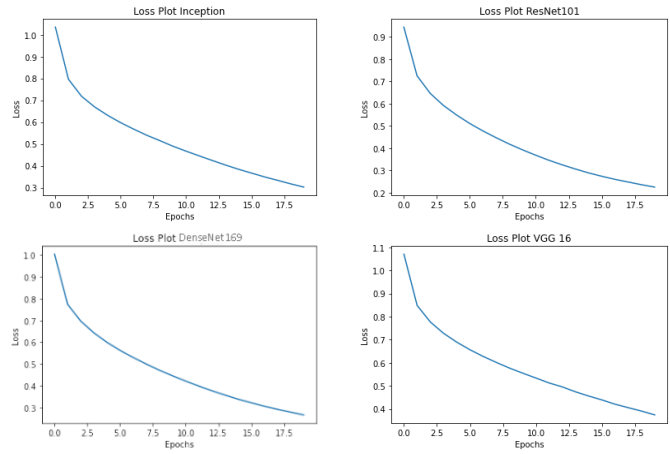

**Fig. 5**   Loss Plot of models pre-trained with different networks

Although the overall results are satisfactory, they can easily be improved by utilizing more data for training, as we only used almost 113,287 images due to computational limitations. Fig.5, illustrates the model's loss plot for four pre-trained networks.

## 4.4   Qualitative results

Our model's qualitative results are pretty intriguing. The model creates relevant and grammatical captions for various images. Fig. 6, depicts some positive outcomes.

**Fig. 6**   Example captions from the conventional approach with our GRU-based encoder-decoder model, as well as their ground truth captions. Ours, it can be shown, may produce more detailed descriptions with more relevant sentence.

While many captions are instructive, some describe a situation that is completely different from the one depicted in the image, while others are completely incomprehensible. Although it is impossible to distinguish between the two tasks performed by the system, we can label the former a failure in image recognition and the latter a failure in text generation. Fig. 7, depicts various examples of such failures.

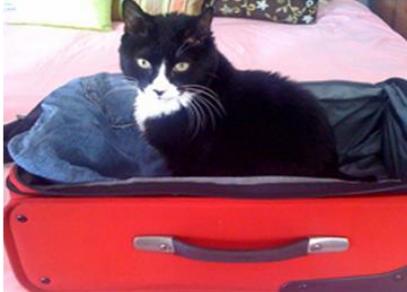
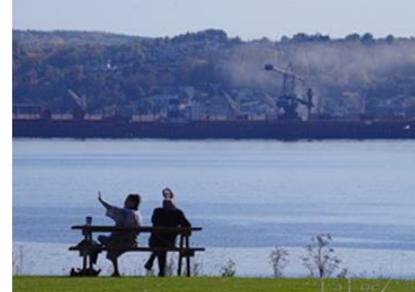
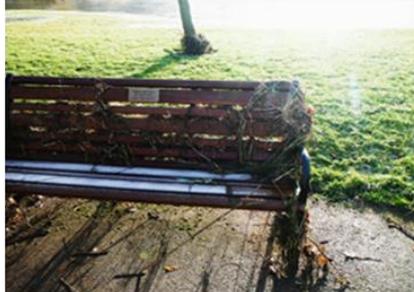

**Fig.7**　　Model predicts poor captions in this case.

Finally, Fig.8, depicts an example output of the attention network architecture. The title is instructive, and the focus on the word teddy bears appears to be in the section of the image with the teddy bear colors.

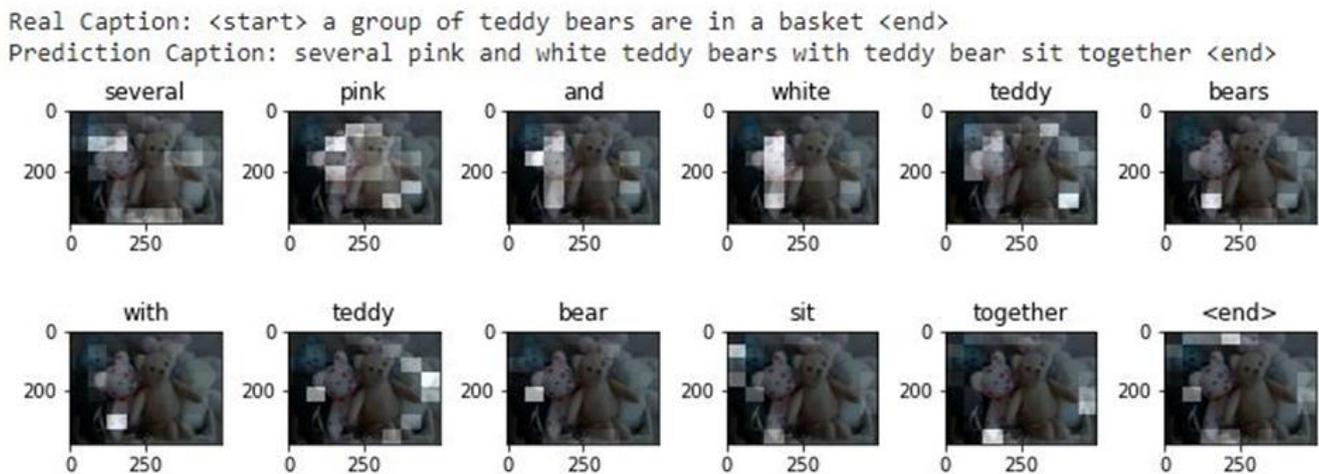

**Fig.8**　　Attention is drawn to different regions of the image in this example. The attention weights in white spots are higher.

## 5. Conclusion

We introduced a single joint model for automatic image captioning based on CNN and GRU with attention network in this research. One encoder-decoder architecture was used in the suggested model. As the encoder, we used various pre-trained convolutional neural networks to encode an image into a compact representation as graphical characteristics. Then, to create the descriptive sentence, a language model called GRU was chosen as the decoder. Meanwhile, we combined the Bahdanau attention model with GRU to allow learning to be focused on a specific portion of the image in order to improve performance. The entire model can be fully trained using stochastic gradient descent, which simplifies the training procedure. Experiments show that the suggested model is capable of automatically generating appropriate captions for images.

**Availability of Data and Systems**

In all of the experiments, the NVIDIA GPU GTX-1070 was employed. This model was created on a GPU with 64GB of memory and an Intel i9 9900k processor. The development process was completed using PyTorch. Dataset collected from the authorized website of MS-COCO. The publication includes the data that was used in this study.

**Conflicts of Interest**

The authors state that there is no conflict of interest in the publication of this paper.

**Acknowledgment**

This research is supported by the Fundamental Research Funds for the Central Universities. (Grant no. WK2350000002).